\RequirePackage{fix-cm}
\documentclass[twocolumn]{svjour3}          
\smartqed  
\makeatletter
\if@twocolumn
  \renewcommand\normalsize{%
   \@setfontsize\normalsize\@xpt{12.5pt}%
   \abovedisplayskip=3 mm plus6pt minus 4pt
   \belowdisplayskip=3 mm plus6pt minus 4pt
   \abovedisplayshortskip=0.0 mm plus6pt
   \belowdisplayshortskip=2 mm plus4pt minus 4pt
   \let\@listi\@listI}%

  \renewcommand\small{%
   \@setfontsize\small{8.5pt}\@xpt
   \abovedisplayskip 8.5\p@ \@plus3\p@ \@minus4\p@
   \abovedisplayshortskip \z@ \@plus2\p@
   \belowdisplayshortskip 4\p@ \@plus2\p@ \@minus2\p@
   \def\@listi{\leftmargin\leftmargini
               \parsep 0\p@ \@plus1\p@ \@minus\p@
               \topsep 4\p@ \@plus2\p@ \@minus4\p@
               \itemsep0\p@}%
   \belowdisplayskip \abovedisplayskip}

\else
  \if@smallext
   \renewcommand\normalsize{%
   \@setfontsize\normalsize\@xpt\@xiipt
   \abovedisplayskip=3 mm plus6pt minus 4pt
   \belowdisplayskip=3 mm plus6pt minus 4pt
   \abovedisplayshortskip=0.0 mm plus6pt
   \belowdisplayshortskip=2 mm plus4pt minus 4pt
   \let\@listi\@listI}%

  \renewcommand\small{%
   \@setfontsize\small\@viiipt{9.5pt}%
   \abovedisplayskip 8.5\p@ \@plus3\p@ \@minus4\p@
   \abovedisplayshortskip \z@ \@plus2\p@
   \belowdisplayshortskip 4\p@ \@plus2\p@ \@minus2\p@
   \def\@listi{\leftmargin\leftmargini
               \parsep 0\p@ \@plus1\p@ \@minus\p@
               \topsep 4\p@ \@plus2\p@ \@minus4\p@
               \itemsep0\p@}%
   \belowdisplayskip \abovedisplayskip}
 \else
  \renewcommand\normalsize{%
   \@setfontsize\normalsize{9.5pt}{11.5pt}%
   \abovedisplayskip=3 mm plus6pt minus 4pt
   \belowdisplayskip=3 mm plus6pt minus 4pt
   \abovedisplayshortskip=0.0 mm plus6pt
   \belowdisplayshortskip=2 mm plus4pt minus 4pt
   \let\@listi\@listI}%

  \renewcommand\small{%
   \@setfontsize\small\@viiipt{9.25pt}%
   \abovedisplayskip 8.5\p@ \@plus3\p@ \@minus4\p@
   \abovedisplayshortskip \z@ \@plus2\p@
   \belowdisplayshortskip 4\p@ \@plus2\p@ \@minus2\p@
   \def\@listi{\leftmargin\leftmargini
               \parsep 0\p@ \@plus1\p@ \@minus\p@
               \topsep 4\p@ \@plus2\p@ \@minus4\p@
               \itemsep0\p@}%
   \belowdisplayskip \abovedisplayskip}
  \fi
\fi

\makeatother

\usepackage[utf8]{inputenc} 

\usepackage[sort,compress]{cite}
\usepackage{graphicx}
\usepackage{amsmath,amssymb}
\usepackage{algorithmic}
\usepackage{array}
\usepackage{url}
\usepackage{color,xspace}
\usepackage[export]{adjustbox}
\usepackage{multirow}
\usepackage{tabularx}
\usepackage{booktabs}
\usepackage{listings}
\usepackage{algorithm,algorithmic}
\usepackage[keeplastbox]{flushend}
\usepackage{xfrac}
\usepackage{hyperref}
\usepackage{makecell}
\usepackage[dvipsnames]{xcolor}
\usepackage{colortbl}
\usepackage{float}
\usepackage{subcaption}

\usepackage{microtype}

\newcommand{\vct}[1]{\ensuremath{\boldsymbol{#1}}} 

\newcommand{\set}[1]{\ensuremath{\mathcal{#1}}}
\newcommand{\con}[1]{\ensuremath{\mathsf{#1}}}
\newcommand{\T}{\ensuremath{\top}}

\newcommand{\argmin}{\operatornamewithlimits{\arg\,\min}}

\newcommand{\tsum}{\textstyle \sum}
\newcommand{\evenn}{\ensuremath{E_1}\xspace}
\newcommand{\evennx}{\ensuremath{{\set E_1}}\xspace}
\newcommand{\evenntwo}{\ensuremath{E_2}\xspace}
\newcommand{\evennxtwo}{\ensuremath{{\set E_2}}\xspace}

\newcommand{\myparagraph}[1]{\smallskip \noindent \textbf{#1}}
\newcommand{\ie}{{i.e.}\xspace}
\newcommand{\eg}{{e.g.}\xspace}

\newcommand{\Drebin}{\textrm{Drebin}\xspace}

\newcommand{\DDrebin}{\textsl{Drebin}\xspace}

\newcommand{\DTesseract}{\textsl{Tesseract}\xspace}

\newcommand{\SVM}{{SVM}\xspace}
\newcommand{\SVMRBF}{{SVM-RBF}\xspace}
\newcommand{\LOGI}{{logistic}\xspace}
\newcommand{\RIDGE}{{ridge}\xspace}
\newcommand{\SECSVM}{{Sec-SVM}\xspace}

\newcommand{\manifest}{\texttt{manifest}\xspace}
\newcommand{\dexcode}{\texttt{dexcode}\xspace}

\newcommand{\grad}{{Gradient}\xspace}
\newcommand{\gradin}{{Gradient*Input}\xspace}
\newcommand{\intgrad}{{Integrated Gradients}\xspace}

\journalname{International Journal of Machine Learning and Cybernetics}

\begin{document}

\title{Do Gradient-based Explanations Tell Anything About Adversarial Robustness to Android Malware?
}


\author{Marco Melis         \and
        Michele Scalas      \and
        Ambra Demontis      \and
        Davide Maiorca      \and
        Battista Biggio    \and
        Giorgio Giacinto    \and
        Fabio Roli
}


\institute{
Corresponding Author: Marco Melis\\
\email{marco.melis@unica.it}\\\\
Marco Melis \and
Ambra Demontis \and Davide Maiorca \and
Battista Biggio \and Giorgio Giacinto \and Fabio Roli \at Department of Electrical and Electronic Engineering, University of Cagliari, Piazza d’Armi 09123, Cagliari, Italy\and
Michele Scalas \and Battista Biggio \and Fabio Roli \at Pluribus One, Italy
}

\date{Received: date / Accepted: date}

\maketitle

\begin{abstract}
While machine-learning algorithms have demonstrated a strong ability in detecting Android malware, they can be evaded by \emph{sparse} evasion attacks crafted by injecting a small set of fake components, \eg, permissions and system calls, without compromising intrusive functionality.
Previous work has shown that, to improve robustness against such attacks, learning algorithms should avoid overemphasizing few discriminant features, providing instead decisions that rely upon a large subset of components.
In this work, we investigate whether gradient-based attribution methods, used to explain classifiers' decisions by identifying the most relevant features, can be used to help identify and select more robust algorithms.
To this end, we propose to exploit two different metrics that represent the \emph{evenness of explanations}, and a new compact security measure called \emph{Adversarial Robustness Metric}. Our experiments conducted on two different datasets and five classification algorithms for Android malware detection show that a strong connection exists between the uniformity of explanations and adversarial robustness.
In particular, we found that popular techniques like \gradin and \intgrad are strongly correlated to security when applied to both linear and nonlinear detectors, while more elementary explanation techniques like the simple \grad do not provide reliable information about the robustness of such classifiers.

\end{abstract}

\section{Introduction}
\label{sect:intro}

Machine learning systems are nowadays being extensively adopted in computer security applications, such as network intrusion and malware detection, as they obtained remarkable performances even against the increasing complexity of modern attacks \cite{AafDuYin13,LinNeuPla15,PenGatSarMol12}. More recently, learning-based techniques based on static analysis proved to be especially effective at detecting Android malware, which constitutes one of the major threats in mobile security. In particular, these approaches showed great accuracy even when traditional code concealing techniques (such as static obfuscation) are employed \cite{rieck14-drebin,chen16-asiaccs,demontis17-tdsc,scalas19-cose,mariconti-ndss17,Chen2021}.

Despite the successful results reported by such approaches, the problem of detecting malware created to fool learning-based systems is still far from being solved. The robustness of machine-learning models is challenged by the creation of the so-called \emph{adversarial examples}, \ie, malicious files that receive fine-grained modifications oriented to deceive the learning-based algorithms \cite{biggio18-pr,biggio13-ecml,szegedy14-iclr,goodfellow15-iclr}. In particular, recent work concerning Android malware demonstrated that specific changes to the contents of malicious Android applications might suffice to change their classification (\eg, from malicious to benign) \cite{demontis17-tdsc,calleja18}, even though the real-word feasibility of these operations should be carefully evaluated \cite{pierazzi2020intriguing, cara2020feasibility}. The main characteristic of these attacks is their \emph{sparsity}, meaning that they enforce only a few changes to the whole feature set to be effective. Such changes may be represented by, \eg, the injection of unused permissions or parts of unreachable/unused executable code. For example, adding a component that is loaded  when the application is started (through a keyword called \texttt{LAUNCHER}) can significantly influence the classifier's decision \cite{melis2018explaining}. 
 
One of the many reasons why such attacks are so effective is that classifiers typically assign significant relevance to a limited amount of features (this phenomenon has also been demonstrated in other applications such as email spam filtering). As a possible countermeasure, research showed that classifiers that avoid overemphasizing specific features, weighting them more evenly, can be more robust against such attacks \cite{kolcz09,biggio10-ijmlc,demontis17-tdsc}. Simple metrics characterizing this behavior were proposed to identify and select more robust algorithms, especially in the context of linear classifiers, where feature weights can be used as a direct measure of a feature's relevance to each decision \cite{demontis16-spr,demontis17-tdsc,demontis19-usenix}. In parallel, the ability to understand the classifiers behavior by looking to the input gradient, \ie the feature weights in the case of linear classifiers, was also explored by multiple works in the field of explainable machine learning  \cite{baehrens10-jmlr,shrikumar2016just,sundararajan2017axiomatic,Adadi2018}. In particular, it became of interest to figure out if the information provided by these gradient-based methods can also be employed to understand (and improve) the robustness of learning-based systems against attacks \cite{chen2019robust}.

In this paper, motivated by the intuition that the classifiers whose attributions are more evenly distributed should also be the more robust, as they rely on a broader set of features for the decision, we propose and empirically validate a few synthetic metrics that allow correlating the \emph{evenness} of gradient-based explanations with the classifier robustness to adversarial attacks. In summary, we make the following contributions:
\begin{itemize}
    \item We statistically investigate the possible correlations between gradient-based explanations, and the classifiers robustness to adversarial \emph{sparse} evasion attacks;
    \item We propose a new measure called \emph{adversarial robustness metric}, to represent the classifier robustness to adversarial attacks along with an increasing attack power in a compact way (Sect.~\ref{sect:evenness});
    \item We assess our findings on the \Drebin \cite{rieck14-drebin} feature space, a popular learning-based detection system for Android, using two different large datasets of applications, \ie, \DDrebin \cite{rieck14-drebin} and \DTesseract \cite{pendlebury2019tesseract, allix2016androzoo}, and five different classification algorithms including linear and non-linear Support Vector Machines, logistic, ridge, and the \emph{secured} linear SVM from \cite{demontis17-tdsc} (Sect.~\ref{sect:exp}).
\end{itemize}

The paper is structured as follows. We first provide a description of learning-based systems for Android malware detection (Sect.~\ref{sect:android}) and their adversarial vulnerabilities (Sect.~\ref{sect:advandroid}). Then, we present the synthetic metrics we use to perform our correlation analysis between the \emph{evenness} of gradient-based explanations and the \emph{adversarial robustness} of classifiers (Sect.~\ref{sect:evenness}). In Sect.~\ref{sect:exp} we present the results of our investigation, which unveils that, under some circumstances, there is a clear relationship between the distribution of gradient-based explanations and the adversarial robustness of Android malware detectors, especially when exploiting more advanced explanation techniques such as \gradin \cite{shrikumar2016just,melis2018explaining} and \intgrad \cite{sundararajan2017axiomatic}. After a brief description of many related works on Android malware detectors, adversarial attacks and explainable machine learning (Sect.~\ref{sect:relwork}), we conclude the paper with a discussion on how our findings can pave the way towards the development of more efficient mechanisms both to evaluate adversarial robustness and to defend against adversarial Android malware examples (Sect.~\ref{sect:conclusions}).

\section{Android Malware Detection}
\label{sect:android}

Here we provide some background on the structure of Android applications, and then we describe \Drebin \cite{rieck14-drebin}, the Android malware detection system used in our analysis.

\subsection{Background on Android}

Android applications are compressed in \texttt{apk} files, \ie, archives that contain the following elements: \emph{(a)} the \texttt{AndroidManifest.xml} file, \emph{(b)} \texttt{classes.dex} files, \emph{(c)} resource and asset files, such as native libraries or images, and \emph{(d)} additional \texttt{xml} files that define the application layout. Since \Drebin analyzes the \texttt{classes.dex} files and the \texttt{AndroidManifest.xml}, we briefly describe them below.

\myparagraph{Android Manifest (\manifest).} The basic information about the Android application is included in the \texttt{AndroidManifest.xml}, including its package name or the supported API levels, together with the declaration of its \emph{components}, \ie, parts of code that perform specific actions. For example, one component might be associated with a screen visualized by the user (\emph{activity}) or to the execution of background tasks (\emph{services}). Application components can also perform actions (through \emph{receivers}) on the occurrence of specific events; for instance, a change in the device's connectivity status (\texttt{CONNECTIVITY\_CHANGE}) or the opening of an application (\texttt{LAUNCHER}). 
The \manifest also contains the list of \emph{hardware components} and \emph{permissions} requested by the app to work (\eg, Internet access).

\myparagraph{Dex bytecode (\dexcode).} The \texttt{classes.dex} file embeds the compiled source code of the applications, including all the user-implemented methods and classes; the bytecode can be executed with the Dalvik Virtual Machine (until Android 4.4) or the Android runtime (ART).
The \texttt{classes.dex} may contain specific API calls that can access sensitive resources such as personal contacts (\emph{suspicious calls}). Additionally, it contains all system-related, \emph{restricted API calls} that require specific permissions (\eg, writing to the device's storage). Finally, this file can contain references to \emph{network addresses} that might be contacted by the application.

\subsection{Drebin}

The majority of the approaches for Android malware detection employ static and dynamic analyses that extract information such as usage of permissions, communications through Inter-Component Communication (ICC), system- and user-implemented API calls, and so forth \cite{rieck14-drebin,lindorfer15-ieee,chen16-asiaccs,scalas19-cose,cai19-tifs}.

\begin{figure*}[t]
\centering
\includegraphics[width=0.95\textwidth]{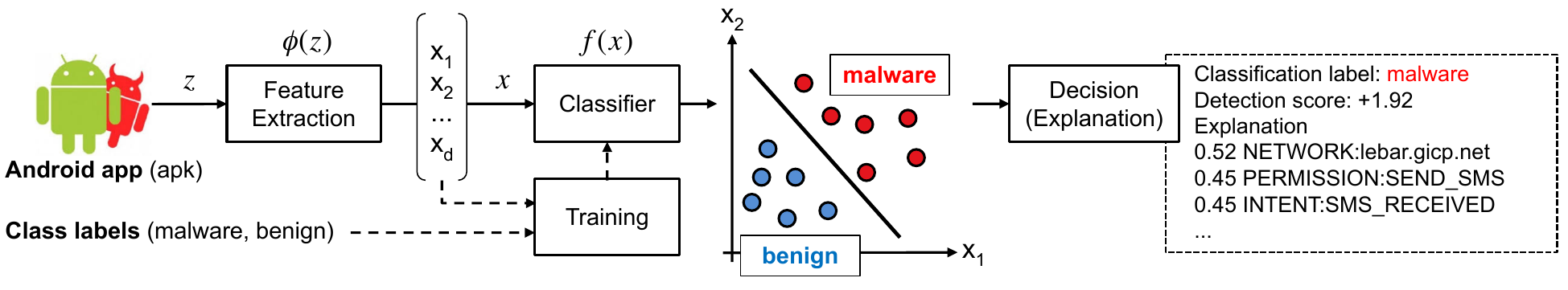}
\caption{A schematic representation (\cite{demontis17-tdsc}) of \Drebin. First, applications are represented as binary vectors in a $\con d$-dimensional feature space. A linear classifier is then trained on an available set of malware and benign applications, assigning a weight to each feature. During classification, unseen applications are scored by the classifier by summing up the weights of the present features: if $f(\vct x) \geq 0$, they are classified as malware. \Drebin also explains each decision by reporting the most suspicious (or benign) features present in the app, along with the weight assigned to them by the linear classifier \cite{rieck14-drebin}}
\label{fig:system-arch}
\end{figure*}

\Drebin is among the most popular and used static detection approaches. It performs the detection of Android malware through static analysis of Android applications. In a first phase (training), it employs a set of benign and malicious apps provided by the user to determine the features that will be used for detection (meaning that the feature set will be strictly dependent on the training data). Such features are then embedded into a \emph{sparse}, high-dimensional vector space. Then, after the training of a linear machine-learning model, the system is able to perform the classification of previously-unseen apps. An overview of the system architecture is given in Fig.~\ref{fig:system-arch}, and discussed more in detail below.

\begin{table}[tp]
	\centering
	\caption{Overview of \Drebin feature sets}
	\label{tab:feature_sets}
    \resizebox{\linewidth}{!}{
	\begin{tabular}{ ll|ll }
		\toprule
		\multicolumn{2}{ c| }{\manifest} & \multicolumn{2}{ c }{\dexcode} \\
		\midrule
		$S_{1}$ & Hardware components & $S_{5}$ & Restricted API calls\\
		$S_{2}$ & Requested permissions & $S_{6}$ & Used permission \\
		$S_{3}$ & Application components & $S_{7}$ & Suspicious API calls \\
		$S_{4}$ & Filtered intents & $S_{8}$ & Network addresses\\
		\bottomrule
	\end{tabular}}
\end{table}

\myparagraph{Feature extraction.} First, \Drebin statically analyzes a set of $\con n$ training Android applications to construct a suitable feature space. All features extracted by \Drebin are presented as \emph{strings} and organized in 8 different feature sets, as listed in Tab.~\ref{tab:feature_sets}.

Android applications are then mapped onto the feature space as follows. Let us assume that an app is represented as an object $\vct z \in \set Z$, being $\set Z$ the abstract space of all \texttt{apk} files. We denote with $\Phi : \set Z \mapsto \set X$ a function that maps an \texttt{apk} file $\vct z$ to a $\con d$-dimensional feature vector $\vct x = ( x^{1}, \ldots, x^{\con d} )^{\T} \in \set X=\{0,1\}^{\con d}$, where each feature is set to 1 (0) if the corresponding \emph{string} is present (absent) in the \texttt{apk} file $\vct z$. An application encoded in feature space may thus look like the following:

{
\centering
    \resizebox{\linewidth}{!}
    {
      \begin{minipage}{\linewidth}
        \centering
        \begin{align}
            \nonumber
            \small \vct x = \Phi( \vct z) \mapsto
            \begin{pmatrix}
            \cdots \\ \small 0\\ \small 1\\
            \cdots \\ \small 1\\ \small 0\\
            \cdots\\
            \end{pmatrix}
            \begin{array}{ll}
            \cdots & \multirow{4}{*}{\hspace{-1mm}\bigg \} $S_2$ }\\
            \texttt{\small permission::SEND\_SMS} \\
            \texttt{\small permission::READ\_SMS}\\
            \cdots & \multirow{4}{*}{\hspace{-1mm}\bigg \} $S_5$ }\\
            \texttt{\small api\_call::getDeviceId}\\
            \texttt{\small api\_call::getSubscriberId}\\
            \cdots & \\
            \end{array}
        \end{align}
      \end{minipage}
    }
}

\myparagraph{Learning and Classification.}
\Drebin uses a linear Support Vector Machine (SVM) to perform detection. It can be expressed in terms of a linear function $f : \set X \mapsto \mathbb R$, \ie, $f(\vct x) = \vct w^{\T}\vct x + b$, where  $\vct w \in \mathbb R^{\con d}$ denotes the vector of \emph{feature weights}, and $b \in \mathbb R$ is the so-called \emph{bias}. These parameters, optimized during training, identify a hyperplane that separates the two classes in the feature space. During classification, unseen apps are then classified as malware if $f(\vct x) \geq 0$, and as benign otherwise.
In this work, we also consider other linear and nonlinear algorithms to learn the classification function $f(\vct x)$.

\myparagraph{Explanation.} \Drebin explains its decisions by reporting, for any given application, the most influential features, \ie, the ones that are present in the given application and are assigned the highest absolute weights by the classifier.
The feature relevance values reported by \Drebin correspond exactly to its feature weights, being \Drebin a linear classifier.
For instance, in Fig.~\ref{fig:system-arch} it is possible to see that \Drebin correctly identifies the sample as malware since it connects to a suspicious URL and uses SMS as a side-channel for communication. In this work, we use different state-of-the-art explainability methods to measure feature relevance and evaluate whether and to which extent the distribution of relevance values reveals any interesting insight on adversarial robustness.

\section{Adversarial Android Malware}
\label{sect:advandroid}
Machine learning algorithms are known to be vulnerable to adversarial examples. The ones used for Android malware detection do not constitute an exception. The vulnerability of those systems was demonstrated in \cite{demontis17-tdsc,grosse17-esorics,demontis19-usenix}, and a defense mechanism was proposed in \cite{demontis17-tdsc}.
In this section, we first explain how an attacker can construct Android malware able to fool a classifier, being recognized as benign. Then, considering the system called Sec-SVM \cite{demontis17-tdsc} as a case-study, we explain how machine learning systems can be strengthened against this attack. 

\subsection{Attacking Android Malware Detection}

The goal of creating adversarial Android malware that evades detection can be formulated as an optimization problem, as detailed below.
This optimization problem is constrained to ensure that the solution provides a functional and realizable malware sample, \ie, that the feature changes suggested by the attack algorithm are feasible and can be implemented as practical manipulations to the actual apk input file.

\myparagraph{Problem Formulation.} As explained in the previous section, Drebin is a binary classifier trained on Boolean features. To have a malware sample $\vct z$ misclassified as benign, the attacker should modify its feature vector $\vct x$ in order to decrease the classifier score $f(\vct x)$. The number of features considered by Drebin is quite large (more than one million). However, the attacker can reasonably change only few of them (\emph{sparse attack}) to preserve the malicious functionality of the application. The attacker has thus an $\ell_1$-norm constraint on the number of features that can be modified. The feature vector of the adversarial application can be computed by solving the following optimization problem:
\begin{align}
\label{eq:evasion}
\argmin_{\vct x^{\prime}}& \quad f(\vct x^{\prime}) \\
\label{eq:evasion-constr}
\rm s. t. & \quad \| \vct x - \vct x^{\prime} \|_1 \leq \varepsilon \\
\label{eq:evasion-box}
& \quad \vct{x}_{\rm lb} \preceq \vct{x}^{\prime} \preceq \vct{x}_{\rm ub}  \\
\label{eq:discrete-constr}
& \quad \vct{x}^{\prime} \in \{0,1\} \enspace ,
\end{align}
where Eq.~\eqref{eq:evasion-constr} is the $\ell_1$ distance constraint between the original $\vct x$ and the modified (adversarial) $\vct x^{\prime}$ sample. Eq.~\eqref{eq:evasion-box} is a box constraint that enforces the features values of the adversarial malware to stay within some lower and upper bounds, while Eq.~\eqref{eq:discrete-constr} enforces the attack to find a Boolean solution.  
The aforementioned problem can be solved with gradient-based optimization techniques, \eg, Projected Gradient Descent (PGD), as described in Alg.~\ref{alg:evasion} \cite{biggio13-ecml,melis17-vipar,demontis19-usenix}. 
At each step, this algorithm projects the feature values of the adversarial sample onto the constraints (Eqs.~\ref{eq:evasion-constr}-\ref{eq:evasion-box}), including binarization in $\{0, 1\}$.

\begin{algorithm}[t]
  \caption{PGD-based attack on Android malware}
  \label{alg:evasion}
  \textbf{Input:} $\vct x$, the input malware; $\varepsilon$, the number of features which can be modified;
  $\eta$, the step size; $\Pi$, a projection operator on the constraints \eqref{eq:evasion-constr} and \eqref{eq:evasion-box}; $t>0$, a small number to ensure convergence.\\
  \textbf{Output:} $\vct x^\prime$, the  adversarial (perturbed) malware.

    \begin{algorithmic}[1]
    \STATE{$\vct x^\prime \gets \vct x$}
    \REPEAT
     \STATE{$\vct x^\star \gets \vct x^\prime$}
     \STATE{$\vct x^\prime \gets \Pi(\vct x^\star - \eta \cdot \nabla f(\vct x^\star))$}
    \UNTIL{$|f(\vct x^\prime) - f(\vct x^\star)| \leq t$}
    \STATE{\textbf{return:} $\vct x^\prime$}
  \end{algorithmic}

\end{algorithm}

\myparagraph{Feature Addition.} To create malware  able to fool the classifier, an attacker may, in theory, both adding and removing features from the original applications. However, in practice, removing features is a non-trivial operation that can easily compromise the malicious functionalities of the application.
Feature addition is a safer operation, especially when the injected features belong to the \manifest; for example, adding permissions does not influence any existing application functionality. When the features depend on the \dexcode, it is possible to add them safely introducing information that is not actively executed, \eg, by adding code after \texttt{return} instructions (\emph{dead code}) or methods that are never called by any \texttt{invoke} type instructions (\ie, the ones that indicate a method call). Therefore, in this work, we only consider feature addition.
To find a solution that does not require removing features from the original application, the attacker can simply define $\vct x^\textrm{lb} = \vct x$ in Eq.~\eqref{eq:evasion-box}.
However, it is worth mentioning that this injection could be easily made ineffective, simply removing all the features extracted from code lines that are never executed. In this way, the attacker is forced to change the executed code, which is more difficult, as it requires considering the following additional and stricter constraints. Firstly, the attacker should avoid breaking the application functionalities. Secondly, they should avoid introducing possible artifacts or undesired functionalities, which may influence the semantics of the original program. Injecting a large number of features may be, therefore, difficult and not always feasible.

\subsection{Sec-SVM: Defending against Adversarial Android Malware}
\label{subsect:secsvm}
In \cite{demontis17-tdsc}, the authors showed that the sparse evasion attack described above is able to fool \Drebin, requiring the injection of a negligible number of features, and they propose a robust counterpart of that classifier. The underlying idea behind their countermeasure is to enforcing the classifier to learn more evenly distribute feature weights since this will require the attacker to manipulating more features to evade the classifier.
To this end, they added a box constraint on the weights $\vct w$ of a linear SVM, obtaining the following learning algorithm (\SECSVM):
\begin{eqnarray}
\label{eq:sec-svm}
\min_{\vct w, b} &&  \tfrac{1}{2}\vct {w}^{\T} \vct w + C \tsum_{i=1}^{\con n} \max \left( 0, 1-y_{i}f(\vct x_{i}) \right) \\
{\rm s. t.} &&  w^{\rm lb}_{k} \leq w_{k} \leq w^{\rm ub}_{k} \, , \, k = 1, \ldots, \con d \enspace , \nonumber
\end{eqnarray}
where the lower and upper bounds on $\vct w$ are defined, respectively, by the vectors $\vct w^{\rm lb} = (w^{\rm lb}_{1}, \ldots, w^{\rm lb}_{\con d})$ and $\vct w^{\rm ub} = (w^{\rm ub}_{1}, \ldots, w^{\rm ub}_{\con d})$, which are application-dependent. Eq.~\eqref{eq:sec-svm} can be easily optimized using a constrained variant of the Stochastic Gradient Descent (SGD) technique, as described in \cite{demontis17-tdsc}.

\section{Do Gradient-based Explanations Help to Understand Adversarial Robustness?}
\label{sect:evenness}

\begin{figure*}[t]
    \centering
    \includegraphics[width=0.9\textwidth]{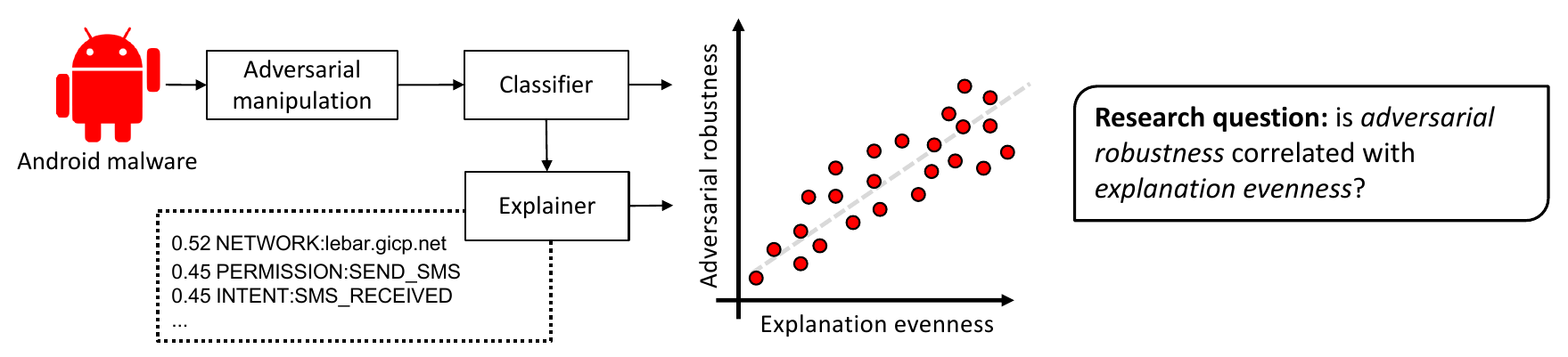}
    \caption{Schematic representation of the analysis employed to verify the correlation between explanation evenness and adversarial robustness. First, for each malware in the test set, we create its adversarial counterpart. Then, for each of those adversarial applications, we evaluate: (1)~a measure of the classifier robustness against it (\emph{adversarial robustness metric}) (2)~the evenness of the application attributions (\emph{explanation evenness}). Finally, we asses the correlation between them}
    \label{fig:evenness-arch}
\end{figure*}

In this work, we investigate whether gradient-based attribution methods used to explain classifiers' decisions provide useful information about the robustness of Android malware detectors against sparse attacks. 
Our intuition is that the classifiers whose attributions are usually evenly-distributed rely upon a broad set of features instead of overemphasizing only a few of them. Therefore, they are more robust against sparse attacks, where the attacker can change only a few features, having a negligible impact on the classifier decision function. 
To verify our intuition, we present an empirical analysis whose procedure is illustrated in Fig.~\ref{fig:evenness-arch} and described below. 
Firstly, we perform a security evaluation on the tested classifier, obtaining a compact measure we call~\emph{Adversarial Robustness Metric} (see Sect.~\ref{sect:adv-robustness}), representing its robustness to the adversarial attacks along with an increasing number of added features $\epsilon$. Then, we compute the attributions for each benign and manipulated malware sample $\vct x$ using a chosen gradient-based explanation technique (see Sect.~\ref{subsect:expltech}) obtaining the relevance vectors $\vct r$. For each of those, we propose to look for a compact metric that encapsulates the degree of~\emph{Evenness} of the attributions (see Sect.~\ref{subsect:evenness-metrics}). Finally, comparing this value with the adversarial robustness metric, we asses the connections between attributions' evenness and the robustness to adversarial evasion attacks. In Sect.~\ref{sect:exp}, we present the results of our analysis on five different learning algorithms trained on the feature space extracted by \Drebin, providing the empirical evidence of our intuition. 

\subsection{Adversarial Robustness Metric}
\label{subsect:adv-robust}

We define the robustness to the evasion samples crafted injecting a fixed number of features $\epsilon$ as:

\label{sect:adv-robustness}
\begin{equation}
R(\set D_{\varepsilon}, f) = \frac{1}{n}\sum_{i=1}^{n} e^{ - \ell_i} \enspace ,
\end{equation}
where $\ell_i = \ell(y_i, f(\vct x_i))$ is the adversarial loss attained by the classifier $f$ on the data points in $\set D_\varepsilon = \{\vct x_i, y_i \}_{i=1}^n$, containing the $\varepsilon$-sized adversarial samples optimized with Algorithm~\ref{alg:evasion}.

Finally, the adversarial robustness metric $\set R$ of a classifier $f$ is defined as the average of $R(\set D_\varepsilon, f)$ on different $\varepsilon$:
\begin{equation}
\set R = \mathbb E_\varepsilon \{ R(\set D_\varepsilon, f) \} \enspace .
\end{equation}

\subsection{Gradient-based Explanation Methods}
\label{subsect:expltech}

In our analysis, we consider gradient-based attribution methods, where \emph{attribution} means the contribution of each input feature to the prediction of a specific sample.
The positive (negative) value of an attribution indicates that the classifier considers the corresponding feature as peculiar of the malicious (benign) samples.
In the following, we review the three gradient-based techniques considered in this work. 

\myparagraph{\grad.} The simplest method to obtain the attributions is to compute the gradient of the discriminant function $f$ with respect to the input sample $\vct x$. For image recognition models, it corresponds to the saliency map of the image \cite{baehrens10-jmlr}.
The attribution of the $i$\textsuperscript{th} feature is computed as:

\begin{align}
    \text{Gradient}_i(\vct x) := \frac{\partial f(\vct x)}{\partial x_i} \enspace .
\end{align}

\myparagraph{\gradin.} This technique has been proposed in \cite{shrikumar2016just} and utilized in one of our previous work \cite{melis2018explaining}, to identify the most influential features for an Android malware detector trained on sparse data. As we have shown in that paper, this approach is more suitable than the previously proposed ones when the feature vectors are sparse. The previously proposed approaches \cite{baehrens10-jmlr,ribeiro16} tended to assign relevance to features whose corresponding components are \emph{not} present in the considered application, thus making the corresponding predictions challenging to interpret. To overcome this issue, this technique leverages the notion of \emph{directional derivative}. Given the input point $\vct x$, it projects the gradient $\nabla f(\vct x)$ onto $\vct x$, to ensure that only the non-null features are considered as relevant for the decision. More formally, the $i$\textsuperscript{th} attribution is computed as:

\begin{align}
    \text{Gradient*Input}_i(\vct x) := \frac{\partial f(\vct x)}{\partial x_i} * x_i \enspace .
\end{align}

\myparagraph{\intgrad.}
Sundararajan et al. \cite{sundararajan2017axiomatic} identified two axioms that attribution methods should satisfy: \emph{implementation invariance}  and  \emph{sensitivity}.
Accordingly to the first, the attributions should always be identical for two functionally equivalent networks, \eg they should be invariant to the differences in the training hyperparameters, which lead the network to learn the same function.
The second axiom is satisfied if, for every input predicted differently from a baseline (a reference vector that models the neutral input, \eg a black image) and that differs from the baseline in only one feature, has, for that feature, a non-zero attribution.
In the same paper, they proposed a gradient-based explanation called Integrated Gradient that satisfies the axioms explained above. This method, firstly, considers the straight-line path from the baseline to the input sample and computes the gradients at all points along the path. Then, it obtains the attribution cumulating those gradients. 
The attribution along the $i$\textsuperscript{th} dimension for an input $\vct x$ and baseline $\vct x^{\prime}$ is defined as:
\begin{equation}
\label{eq:integrads}
\begin{aligned}
&\text{IntegratedGrads}_{i}(\vct x) :=\\&\qquad\left(x_{i}-x_{i}^{\prime}\right) \cdot \int_{\alpha=0}^{1} \frac{\partial f\left(\vct x^{\prime}+\alpha \cdot\left(\vct x-\vct x^{\prime}\right)\right)}{\partial x_{i}} d \alpha \enspace .
\end{aligned}
\end{equation}
To efficiently approximate the previous integral, one can sum the gradients computed at $p$ fixed intervals along the joining path from $\vct x^{\prime}$ to the input $\vct x$:
\begin{equation}
\label{eq:integrads-approx}
\begin{aligned}
&\text{IntegratedGrads}_{i}^{\text{approx}}(\vct x) :=\\&\qquad\left(x_{i}-x_{i}^{\prime}\right) \cdot \sum_{k=1}^{ p} \frac{\partial f\left(\vct x^{\prime}+\frac{k}{ p} \cdot\left(\vct x-\vct x^{\prime}\right)\right)}{\partial x_{i}} \cdot \frac{1}{ p} \enspace .
\end{aligned}
\end{equation}
For linear classifiers, where $\partial f / \partial x_{i} = w_i$, this method is equivalent to Gradient*Input if $\vct x^{\prime} = \vct 0$ is used as a baseline, which is a well-suited choice in many applications \cite{sundararajan2017axiomatic}. Therefore, in this particular case, also the Gradient*Input method satisfies the abovementioned axioms. 

\subsection{Explanation Evenness Metrics}
\label{subsect:evenness-metrics}
To compute the evenness of the attributions, we consider the two metrics, described below.
The first is the one proposed in \cite{kolcz09,biggio10-ijmlc}.  
To compute the evenness metric, they firstly defined a function
$F(\vct r, k)$ which, given a relevance vector $\vct r$, computes the ratio of the sum of the k highest relevance values to
the sum of all absolute relevance values, for $k = 1,2,\ldots,m$:
\begin{center}
\begin{align}
    F(\vct r,k) = \frac{\sum_{i=1}^k |r_{(i)}|}{\sum_{j=1}^{ m} |r_{(j)}|} \enspace ,
\nonumber
\end{align}
\end{center}
where $r_{1}, r_{2}, \ldots, r_{ m}$ denote the relevance values, sorted in descending order of their absolute values, \ie, $|r_{1}| \geq |r_{2}| \geq \ldots \geq |r_{ m}|$ and $m$ is the number of considered relevance values ($m\leq d$).
This function essentially computes the evenness of the distribution of the relevance among the features. 
The evenest relevance distribution (the one where they are all equal), corresponds to $F(\vct r, k) = k/n$. Whereas the most uneven is attained when only one relevance differs from zero, and in this case, $F(\vct r, k) = 1$ for each k value. To avoid the dependence on $k$ and to obtain a single scalar value, they compute the evenness as:
\begin{align}
    \evennx(\vct r) = \frac{2}{ m - 1}\left[ m - \sum_{k=1}^{ m} F(\vct r, k)\right] \enspace .
\label{eq:rel-evenn}
\end{align}
The range of $\evennx$ is $[0, 1]$, $\evennx = 0$ and $\evennx = 1$ indicates respectively to the most uneven and to the most even relevance vector.

The second metric we consider is the one proposed in \cite{demontis16-spr}, based on the ratio between the $\ell_1$ and $\ell_\infty$ norm:
\begin{align}
    \evennxtwo(\vct r) = \frac{1}{m} \cdot \frac{\|\vct r\|_1}{\|\vct r\|_\infty} \enspace .
\label{eq:rel-evenn-spr}
\end{align}

To have a broader perspective of the attributions' evenness, we compute the metrics on multiple samples, and we average the results. More formally, we define the \emph{explanation evenness} as:
\begin{align}
    E=\frac{1}{n} \sum_{i=1}^n \set E (\vct r^i) \enspace ,
\label{eq:rel-evenn-global}
\end{align}
where $\vct r^i$ with $i=1, 2, \ldots, n$ is the attribution vector computed on each sample of a test dataset $\set D = \{\vct x_i, y_i \}_{i=1}^n$, and $\set E$ can be equal either to $\evennx$ or $\evennxtwo$.
In the following, we represent the averaged evenness computed considering the per-sample metric $\evennx$ ($\evennxtwo$) with \evenn (\evenntwo).

\subsection{Computational Complexity}
An important aspect to consider when performing the adversarial robustness evaluation of a classifier is the complexity of the process. Especially in the case of non-linear classifiers, computing a full security evaluation curve for this purpose may require hundred thousands iterations~\cite{biggio18-pr}. 
Because, for each test sample, the corresponding adversarial example should be computed, this process involves, for every single point, thousand of iterations (and thus of gradient and function evaluations) with the chosen optimization algorithm\cite{tramer20-nips}.
In this sense, exploiting the gradient-based attributions methods provides instead a massive computational advantage. In fact, for both \grad and \gradin methods, only a single gradient evaluation is required to obtain the attributions of an input sample, and this gradient is even identical for all samples in the case of linear classifiers (allowing to save even more evaluations). For the \intgrad technique, the number of gradient evaluations depends on the chosen value of the $p$ parameter in Eq.~\eqref{eq:integrads-approx} which, however, is usually set to a small number. In our correlation analysis, we assume that the attack samples required to compute the adversarial robustness metric $\set R$ and the attributions for each of the test samples are already acquired as part of the security and explainability evaluation of the classifiers. The complexity of the process is then given by the computation of the explanation evenness and the adversarial robustness metric.

\section{Experimental Analysis} 
\label{sect:exp}

In this section, we practically evaluate whether the measures introduced in Sect.~\ref{sect:evenness} can be used to estimate the robustness of classifiers against sparse evasion attacks.
After detailing our experimental setup (Sect.~\ref{subsect:setup}), we show the classifiers' detection performances, both in normal conditions and under attack (Sect.~\ref{subsect:perfres}).
In our evaluations, we focus on the feature addition attack setting (see Sect.~\ref{sect:advandroid}), as they are typically the easiest to accomplish for the adversary. We use \texttt{secml} as a framework to implement classification systems, explanation techniques, and attack algorithms \cite{melis2019secml}.
Finally, we assess the relationship of the proposed evenness metrics with our new adversarial robustness metric and the detection rate (Sect.~\ref{subsect:correlsres}).

\subsection{Experimental Setup}
\label{subsect:setup}

\begin{figure}[t]
\centering
\includegraphics[width=0.475\textwidth]{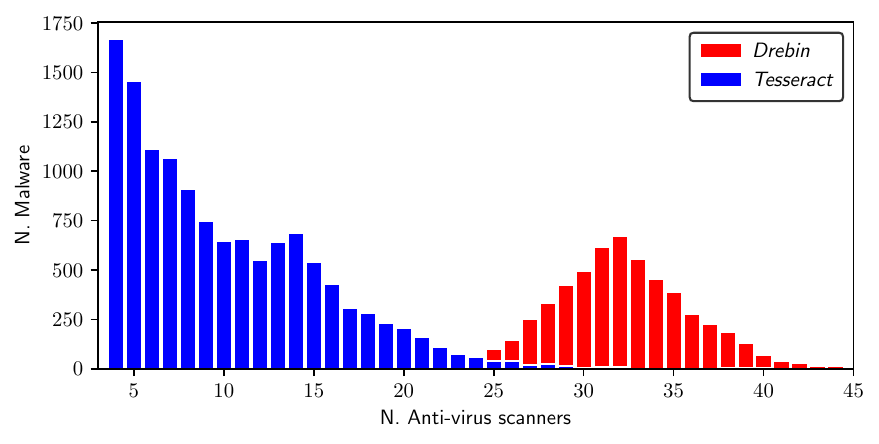}
\caption{Distribution of malware in each dataset with respect to the number of anti-virus scanners that flagged the applications as positive. 
Data extracted from \texttt{VirusTotal}
}
\label{fig:vt-counts}
\end{figure}

\myparagraph{Datasets.} We use two different datasets of real-world Android applications. The first is the \DDrebin dataset \cite{rieck14-drebin}, consisting of $121,329$ benign applications and $5,615$ malicious samples, labeled with \texttt{VirusTotal} and collected between August 2010 and October 2012. A sample is labeled as malicious if it is flagged by at least five anti-virus scanners, whereas it is labeled as benign otherwise. The second is the \DTesseract dataset \cite{pendlebury2019tesseract}, consisting of $116,993$ benign applications and $12,735$ malicious samples, collected from \texttt{AndroZoo} \cite{allix2016androzoo} between January 2014 and December 2016. In Fig.~\ref{fig:vt-counts} we report the distribution of malware in each dataset with respect to the number of anti-virus scanners that flagged the applications as positive. We can observe how for \DDrebin most samples are flagged by 30 to 35 scanners, while for \DTesseract 4 to 10 scanners detect most of the positives. This shows how recognizing the samples of the latter, newer dataset, still represents a significant challenge for many scanners.

\myparagraph{Training-validation-test splits.} We average our results on 5 runs. In each run, we randomly selected 60,000 apps from both datasets to train the learning algorithms, and we used the remaining apps for testing.

\myparagraph{Classifiers.} We compare the standard \DDrebin implementation based on a linear Support Vector Machine (\SVM) against the \textit{secured} linear SVM from \cite{demontis17-tdsc} (\SECSVM), an SVM with the RBF kernel (\SVMRBF), a logistic regression (\LOGI) and a ridge regression (\RIDGE). 

\myparagraph{Parameter setting.} Using a 10-fold cross-validation procedure, we optimize the parameters of each classifier to maximize the detection rate (\ie, the fraction of detected malware) at $1\% $ false-positive rate (\ie, the fraction of legitimate applications misclassified as malware).
In particular, we optimize the parameters $C \in \{10^{-2}, 10^{-1}, \ldots, 10^{2}\}$ for \LOGI and both linear and non-linear SVMs, the kernel $\gamma \in \{10^{-4}, 10^{-3}, \ldots, 10^{2}\}$ for the \SVMRBF, and $\alpha \in \{10^{-2}, 10^{-1}, \ldots, 10^{2}\}$ for \RIDGE. For \SECSVM, we optimized the $-\vct w^{\rm lb} = \vct w^{\rm ub} \in \{0.1, 0.25, 0.5\}$ and $C \in \{10^{-2}, 10^{-1}, \ldots, 10^{2}\}$.
When similar detection rates ($\pm 1\%$) are obtained for different hyperparameter configurations, we select the configuration corresponding to a more regularized classifier, as more regularized classifiers are expected to be more robust under attack \cite{demontis19-usenix}.  
The typical values of the aforementioned hyperparameters found for both datasets after cross-validation are $C=0.1$ for \SVM, $\alpha = 10$ for \RIDGE, $C=1$ for \LOGI, $C=1$ and $w=0.25$ for \SECSVM, $C=10$ and $\gamma=0.01$ for \SVMRBF. 

\myparagraph{Attribution computation} For each dataset, we compute the attributions on $1,000$ malware samples randomly chosen from the test set. We took $\vct x^{\prime} = 0$ as the baseline for \intgrad, and we compute the attributions with respect to the malware class. As a result, positive (negative) relevance values in our analysis denote malicious (benign) behavior.
Given the high sparsity ratio of the feature space, we use $m = 1,000$ to compute the explanation evenness metrics.

\subsection{Experimental Results}
\label{subsect:perfres}

\begin{figure*}[t]
\centering
\begin{subfigure}[t]{\textwidth}
\centering
\includegraphics[width=.445\textwidth,trim={0 0 3cm 0},clip]{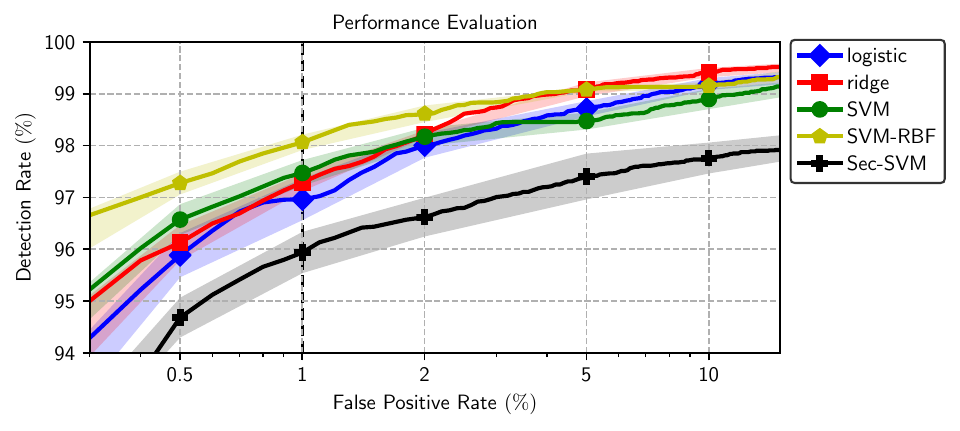}
\includegraphics[width=.523\textwidth,trim={.7cm 0 0 0},clip]{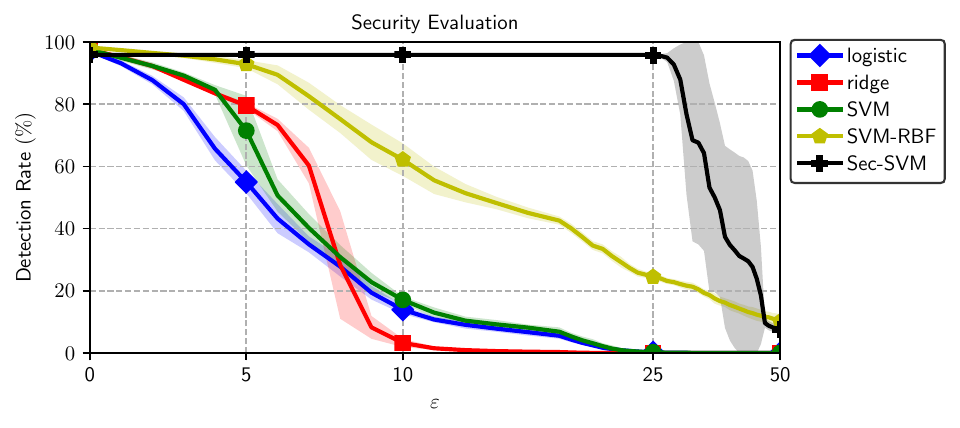}
\caption{\DDrebin}
\label{fig:res-drebin}
\vspace{1em}
\end{subfigure}
\begin{subfigure}[t]{\textwidth}
\centering
\includegraphics[width=.445\textwidth,trim={0 0 3cm 0},clip]{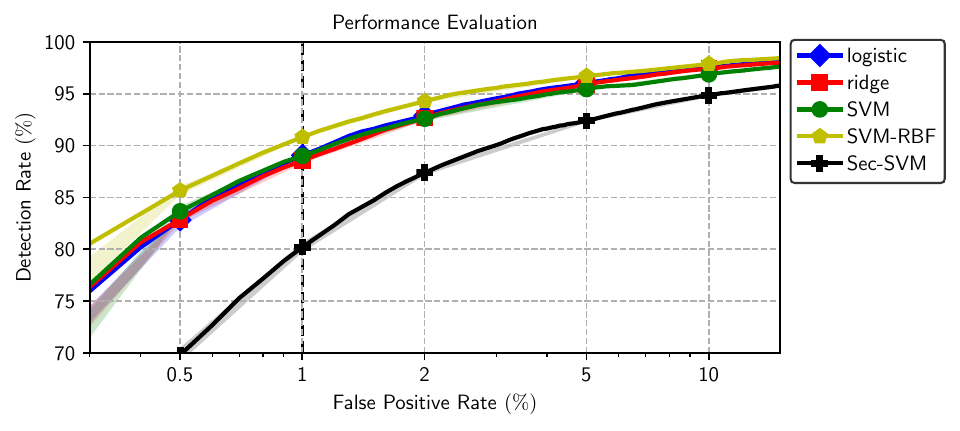}
\includegraphics[width=.523\textwidth,trim={.7cm 0 0 0},clip]{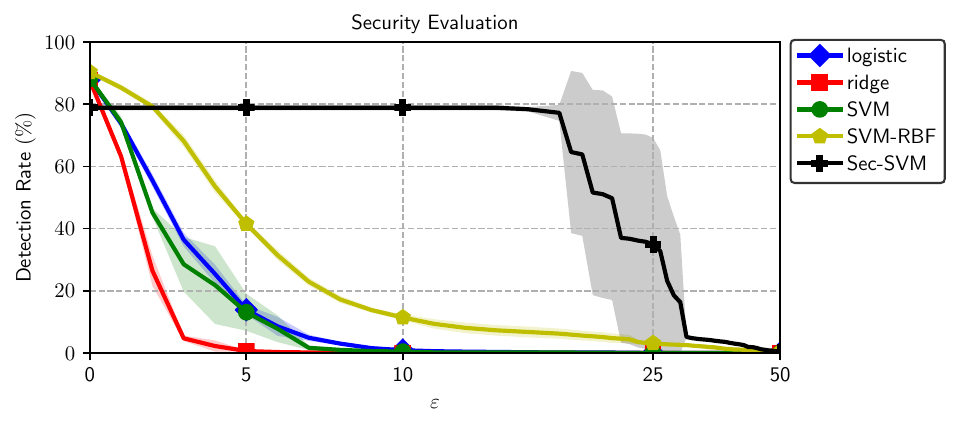}
\caption{\DTesseract}
\label{fig:res-parrot}
\end{subfigure}
\caption{(left) Mean ROC curves for the tested classifiers. (right) White-box evasion attacks. Detection Rate at 1\% False Positive Rate against an increasing number of added features $\varepsilon$. We can see how the Sec-SVM, despite providing a slightly lower detection rate compared to the other tested classifiers on both datasets, requires on average more than 20 different new feature additions to the original applications to be fooled by the attacker}
\label{fig:res}
\end{figure*}

We first perform an evaluation of the performances under normal conditions; the resulting Receiver Operating Characteristic (ROC) curves with the Detection Rate for each classifier, averaged over the 5 repetitions, is reported in the left side of Fig.~\ref{fig:res-drebin} and Fig.~\ref{fig:res-parrot}.
We then perform a white-box evasive attack against each classifier, aiming to have $1000$ malware samples randomly chosen from the test sets misclassified as benign. The results are shown on the right side of Fig.~\ref{fig:res-drebin} and Fig.~\ref{fig:res-parrot}, which report the variation of the detection rate as the number of modified features $\varepsilon$ increases. On both datasets, we can notice how the \SECSVM classifier (described in Sect.~\ref{subsect:secsvm}) provides a slightly worse detection rate compared to the other classifiers, but is particularly robust against adversarial evasion attacks.

\subsection{Is adversarial robustness correlated with explanation evenness?} 
\label{subsect:correlsres}

\begin{table}[t]
\centering
\begin{adjustbox}{width=.24\textwidth,valign=t}
\begin{tabular}{cp{5.45cm}r}
\multicolumn{3}{c}{
\SVMRBF ($\evennx = 46.24\%$, $\evennxtwo = 22.47\%$, $\varepsilon_{\rm min} = 6$)}\\
\toprule
   \textbf{Set} & \textbf{Feature Name} & \multicolumn{1}{c}{$\vct r$ (\%)} \\
\midrule
          S2 & \cellcolor{red!50} SEND\_SMS \rule{0pt}{9pt} & 10.35 \\
     S7 & \cellcolor{red!50} \makecell[tl]{android/telephony/TelephonyManager\\;-\ensuremath{>}getNetworkOperator} & 10.05 \\
     S4 & \cellcolor{NavyBlue!40} LAUNCHER                          & -8.89 \\
     S5 & \cellcolor{NavyBlue!40} \makecell[tl]{android/os/PowerManager\$WakeLock\\;-\ensuremath{>}release}                     & -8.01 \\
     S2 & \cellcolor{red!30} READ\_PHONE\_STATE                      &  5.03 \\
     S2 & \cellcolor{NavyBlue!30} RECEIVE\_SMS                           & -5.00 \\
     S3 & \cellcolor{red!20} c2dm.C2DMBroadcastReceiver                  &  4.56 \\
     S2 & \cellcolor{red!20} READ\_SMS                              &  3.52 \\
     S4 & \cellcolor{red!20} DATA\_SMS\_RECEIVED                   &  3.50 \\
     S5 & \cellcolor{NavyBlue!20} \makecell[tl]{android/app/NotificationManager\\;-\ensuremath{>}notify}                       & -3.49 \\
\bottomrule
\end{tabular}
\end{adjustbox}%
\begin{adjustbox}{width=.24\textwidth,valign=t}
\begin{tabular}{cp{5.45cm}r}
\multicolumn{3}{c}{
\SECSVM ($\evennx = 73.04\%$, $\evennxtwo = 66.24\%$, $\varepsilon_{\rm min} = 31$)}\\
\toprule
   \textbf{Set} & \textbf{Feature Name} & \multicolumn{1}{c}{$\vct r$ (\%)} \\
\midrule
          S2 & \cellcolor{red!20} READ\_PHONE\_STATE \rule{0pt}{10pt} & 3.51 \\
     S7 & \cellcolor{red!20} \makecell[tl]{android/telephony/TelephonyManager\\;-\ensuremath{>}getNetworkOperator} & 3.51 \\
     S2 & \cellcolor{red!20} SEND\_SMS  \rule{0pt}{9pt} & 3.51 \\
     S3 & \cellcolor{red!20} c2dm.C2DMBroadcastReceiver  \rule{0pt}{11pt} & 3.51 \\
     S2 & \cellcolor{red!20} INTERNET\rule{0pt}{10pt} & 3.44 \\
     S3 & \cellcolor{red!20} com.software.application.ShowLink \rule{0pt}{9pt}  & 3.39 \\
     S3 & \cellcolor{red!20} com.software.application.Main \rule{0pt}{9pt} & 3.39 \\
     S3 & \cellcolor{red!20} com.software.application.Notificator \rule{0pt}{9pt}  & 3.39 \\
     S3 & \cellcolor{red!20} com.software.application.Checker  \rule{0pt}{9pt} & 3.39 \\
     S3 & \cellcolor{red!20} com.software.application.OffertActivity \rule{0pt}{9pt} & 3.39 \\
\bottomrule
\end{tabular}
\end{adjustbox}\\\vspace{.5em}%
\begin{adjustbox}{width=.24\textwidth,valign=t}
\begin{tabular}{cp{5.45cm}r}
\multicolumn{3}{c}{
\SVMRBF ($\evennx = 60.74\%$, $\evennxtwo = 25.84\%$, $\varepsilon_{\rm min} = 31$)}\\
\toprule
   \textbf{Set} & \textbf{Feature Name} & \multicolumn{1}{c}{$\vct r$ (\%)} \\
\midrule
     S4 & \cellcolor{NavyBlue!10} LAUNCHER \rule{0pt}{8pt} & -1.89 \\
     S7 & \cellcolor{red!10} android/net/Uri;-\ensuremath{>}fromFile \rule{0pt}{8pt} &  1.34 \\
     S5 & \cellcolor{NavyBlue!10} \makecell[tl]{android/os/PowerManager\$WakeLock\\;-\ensuremath{>}release}                  & -1.25 \\
     S2 & \cellcolor{red!10} INSTALL\_SHORTCUT \rule{0pt}{8pt} &  1.23 \\
     S7 & \cellcolor{NavyBlue!10} \makecell[tl]{android/telephony/SmsMessage\\;-\ensuremath{>}getDisplayMessageBody} & -1.21 \\
     S7 & \cellcolor{NavyBlue!10} \makecell[tl]{android/telephony/SmsMessage\\;-\ensuremath{>}getTimestampMillis}    & -1.20 \\
     S2 & \cellcolor{NavyBlue!10} SET\_ORIENTATION \rule{0pt}{8pt} & -1.20 \\
     S2 & \cellcolor{red!10} ACCESS\_WIFI\_STATE \rule{0pt}{8pt} &  1.15 \\
     S4 & \cellcolor{red!10} BOOT\_COMPLETED \rule{0pt}{8pt}&  1.08 \\
     S5 & \cellcolor{NavyBlue!10} android/media/MediaPlayer;-\ensuremath{>}start \rule{0pt}{8pt} & -1.06 \\
\bottomrule
\end{tabular}
\end{adjustbox}%
\begin{adjustbox}{width=.24\textwidth,valign=t}
\begin{tabular}{cp{5.45cm}r}
\multicolumn{3}{c}{
\SECSVM ($\evennx = 63.14\%$, $\evennxtwo = 52.70 \%$, $\varepsilon_{\rm min} = 39$)}\\
\toprule
   \textbf{Set} & \textbf{Feature Name} & \multicolumn{1}{c}{$\vct r$ (\%)} \\
\midrule
     S2 & \cellcolor{red!5} ACCESS\_NETWORK\_STATE                      &  0.93 \\
     S2 & \cellcolor{red!5} READ\_PHONE\_STATE                          &  0.93 \\
     S6 & \cellcolor{red!5} READ\_HISTORY\_BOOKMARKS                                      &  0.93 \\
     S7 & \cellcolor{NavyBlue!5} \makecell[tl]{android/telephony/TelephonyManager\\;-\ensuremath{>}getNetworkOperatorName} & -0.93 \\
     S6 & \cellcolor{NavyBlue!5} ACCESS\_NETWORK\_STATE                                        & -0.93 \\
     S7 & \cellcolor{red!5} android/telephony/SmsMessage;-\ensuremath{>}getDisplayOriginatingAddress &  0.93 \\
     S7 & \cellcolor{red!5} \makecell[tl]{android/telephony/TelephonyManager\\;-\ensuremath{>}getNetworkOperator}     \rule{0pt}{9pt} &  0.93 \\
     S7 & \cellcolor{NavyBlue!5} android/net/Uri;-\ensuremath{>}getEncodedPath                           \rule{0pt}{9pt} & -0.93 \\
     S2 & \cellcolor{NavyBlue!5} SET\_ORIENTATION                         \rule{0pt}{9pt}  & -0.93 \\
     S7 & \cellcolor{red!5} java/lang/reflect/Method;-\ensuremath{>}invoke                         \rule{0pt}{8pt}  &  0.93 \\
\bottomrule
\end{tabular}
\end{adjustbox}
\caption{Top-10 influential features and corresponding \gradin relevance ($\%$) for a malware of the \texttt{FakeInstaller} family (top) and a malware of the \texttt{Plankton} family (bottom). Notice that the minimum number of features to add $\varepsilon_\text{min}$ to evade the classifiers increases with the evenness metrics \evennx and \evennxtwo}
\label{tab:local-ranks}
\end{table}

We now investigate the connection between adversarial robustness and evenness of gradient-based explanations. We start with two illustrative examples.
Tab.~\ref{tab:local-ranks} shows the top-10 influential features for two malware samples from the \DDrebin dataset, one of \texttt{FakeInstaller}\footnote{MD5: f8bcbd48f44ce973036fac0bce68a5d5} and one of \texttt{Plankton}\footnote{MD5: eb1f454ea622a8d2713918b590241a7e} family, reported for the \SVMRBF and \SECSVM algorithms, and obtained through the \gradin technique. All the classifiers correctly label the samples as malware. 

Looking at the features of the \texttt{FakeInstaller} malware, we can observe how both the classifiers identify the cellular- and SMS-related features, \eg, the \texttt{GetNetworkOperator()} method or the \texttt{SEND\_SMS} permission, as highly relevant. This is coherent with the actual behavior of the malware sample since its goal is to send SMS messages to premium-rate numbers. 
With respect to the relevance values, the first aspect to point out comes from their relative magnitude, expressed as a percentage in Tab.~\ref{tab:local-ranks}. In particular, we can observe that the top-10 relevance values for \SVMRBF vary, regardless of their signs, from $3.49\%$ to $10.35\%$, while for \SECSVM the top values lie in the $3.39\%$--$3.51\%$ range. This suggests that \SVMRBF assigned high prominence to few features; conversely, \SECSVM distributed the relevance values more evenly. It is possible to catch this behavior more easily through the synthetic evenness measures $\evennx$ (Eq.~\eqref{eq:rel-evenn}) and $\evennxtwo$ (Eq.~\eqref{eq:rel-evenn-spr}) reported in Tab.~\ref{tab:local-ranks}, which show higher values for \SECSVM. 
Tab.~\ref{tab:local-ranks} also shows the $\varepsilon_\text{min}$ value, \ie, the minimum number of features to add to the malware to evade the classifier. We can notice how the $\varepsilon_\text{min}$ parameter is strictly related to the evenness distribution, since higher values of $\evennx$ and $\evennxtwo$ correspond to higher values of $\varepsilon_\text{min}$, \ie, a higher effort for the attacker to accomplish her goal. In particular, it is possible to identify a clear difference between the behavior of \SVMRBF and \SECSVM: the diversity of their evenness metrics, which cause the $\varepsilon_\text{min}$ values to be quite different as well, indicates that, for this prediction, \SVMRBF is quite susceptible to a possible attack compared to \SECSVM.

Conversely, considering the second sample, the attributions (regardless of the sign) and the evenness metrics present similar values. Such behavior is also reflected in the associated $\varepsilon_\text{min}$ values. In this case, the relevance values are more evenly distributed, which indicates that the evasion is more difficult.

\begin{figure*}[t]
\begin{subfigure}[t]{0.49\textwidth}
\centering
     \includegraphics[width=0.5\textwidth,trim={0 .7cm 0 0},clip]{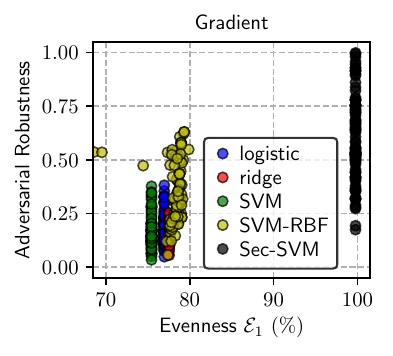}
    \includegraphics[width=0.448\textwidth,trim={.7cm .7cm 0 0},clip]{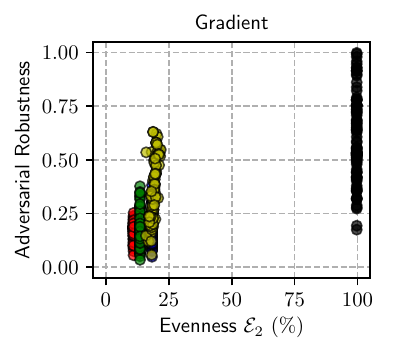}\\
    \includegraphics[width=0.5\textwidth,trim={0 .7cm 0 0},clip]{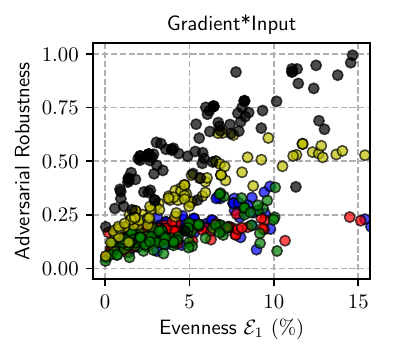}
    \includegraphics[width=0.448\textwidth,trim={.7cm .7cm 0 0},clip]{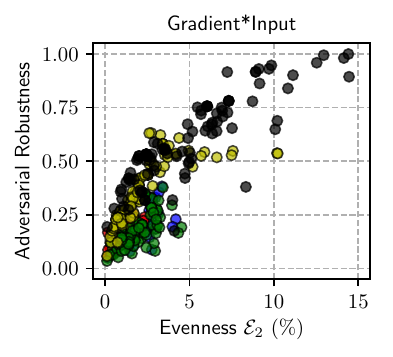}\\
    \includegraphics[width=0.5\textwidth,trim={0 0 0 0},clip]{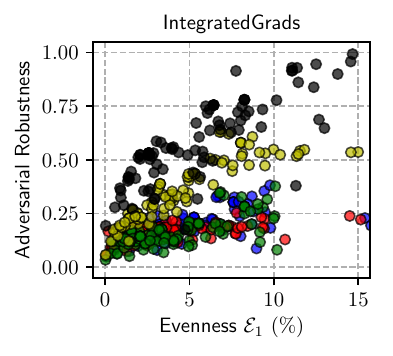}
    \includegraphics[width=0.448\textwidth,trim={.7cm 0 0 0},clip]{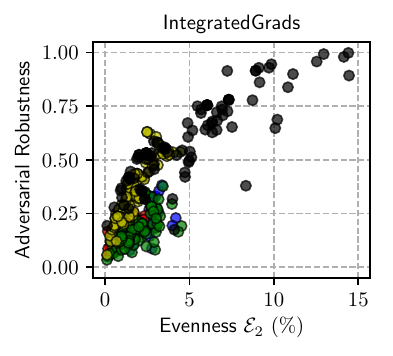}
    \caption{\DDrebin}
    \label{fig:mscore_scatter_drebin}
    \end{subfigure}
    \hfill
\begin{subfigure}[t]{0.49\textwidth}
\centering
     \includegraphics[width=0.5\textwidth,trim={0 .7cm 0 0},clip]{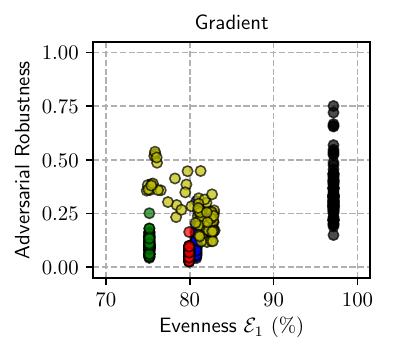}
    \includegraphics[width=0.448\textwidth,trim={.7cm .7cm 0 0},clip]{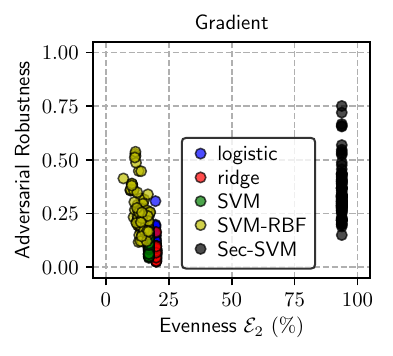}\\
    \includegraphics[width=0.5\textwidth,trim={0 .7cm 0 0},clip]{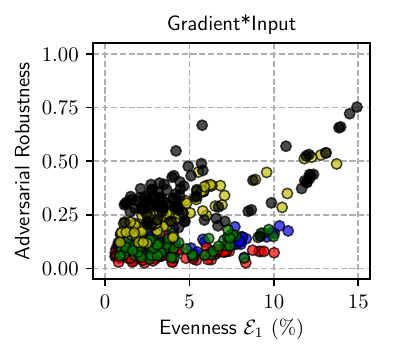}
    \includegraphics[width=0.448\textwidth,trim={.7cm .7cm 0 0},clip]{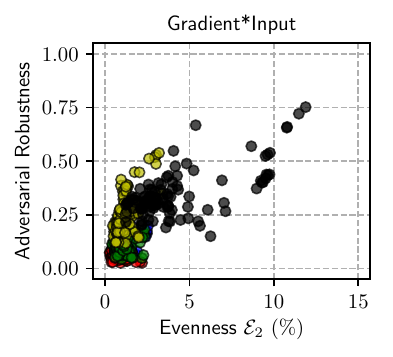}\\
    \includegraphics[width=0.5\textwidth,trim={0 0 0 0},clip]{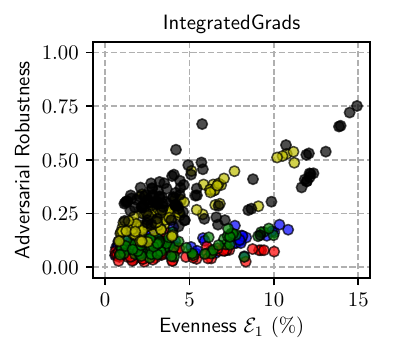}
    \includegraphics[width=0.448\textwidth,trim={.7cm 0 0 0},clip]{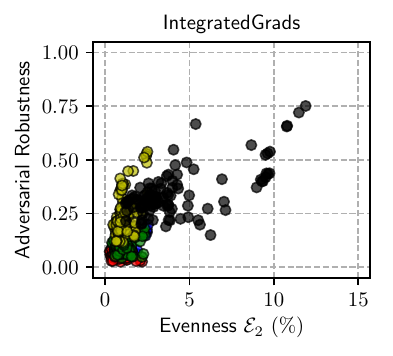}
    \caption{\DTesseract}
    \label{fig:mscore_scatter_parrot}
    \end{subfigure}
    \caption{Evaluation of the adversarial robustness metric $\set R$ against the evenness \evennx, \evennxtwo metrics for the different gradient-based explanation techniques computed on $1000$ samples of the test set (only $100$ samples are shown)}
    \label{fig:mscore_scatter}
\end{figure*}

We now correlate the evenness metrics with the \emph{adversarial robustness metric} $\set R$, introduced in Sect.~\ref{subsect:adv-robust}.
Fig.~\ref{fig:mscore_scatter} shows the relationship between this value and the evenness metrics for $100$ samples chosen from the test set of \DDrebin (Fig.~\ref{fig:mscore_scatter_drebin}) and \DTesseract (Fig.~\ref{fig:mscore_scatter_parrot}), reported for each explainability technique. From this broader view, we can see how the evenness values calculated on top of the \gradin and \intgrad explanations present a significant connection to the adversarial robustness metric for both datasets. 
This seems not applicable to the \grad technique, which appears to be weakly correlated with explanation evenness. Specifically, we observe in Fig.~\ref{fig:mscore_scatter} that the dots of the linear classifiers are perfectly vertical-aligned. This fact is caused by the constant value of the gradient across all the samples, which implies constant values for the evenness metrics as well. The reliability of this technique appears to be low even in the case of SVM-RBF, especially for the \DTesseract dataset where we observe a negative correlation with the explanation evenness.

In order to assess the statistical significance of these plots, we also compute the associated correlation values with three different metrics: Pearson (P), Spearman Rank (S), Kendall's Tau (K). The results are shown in Tab.~\ref{tab:mscore_correlation_drebin} and Tab.~\ref{tab:mscore_correlation_parrot}. In the case of \DDrebin data, we obtain a strong $p$-val $\ll 0.05$ for all the tested classifiers using both \gradin and \intgrad, confirming the validity of our findings from Fig.~\ref{fig:mscore_scatter}. The same is valid in the case of \DTesseract data, with $p$-val $< 0.01$ for the two explanation techniques and both evenness metrics in all cases.

\begin{figure*}[t]
\begin{subfigure}[t]{0.49\textwidth}
\centering
    \includegraphics[width=0.5\textwidth,trim={0 0 0 0},clip]{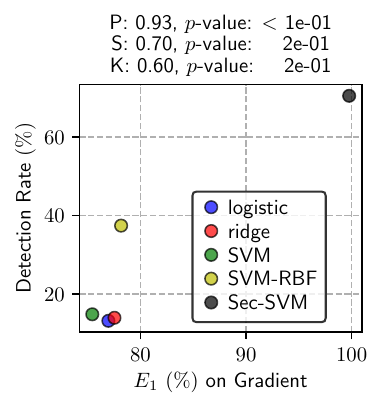}
    \includegraphics[width=0.446\textwidth,trim={.7cm 0 0 0},clip]{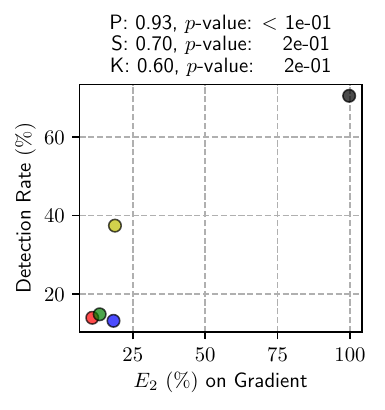}\\
    \includegraphics[width=0.5\textwidth,trim={0 0 0 0},clip]{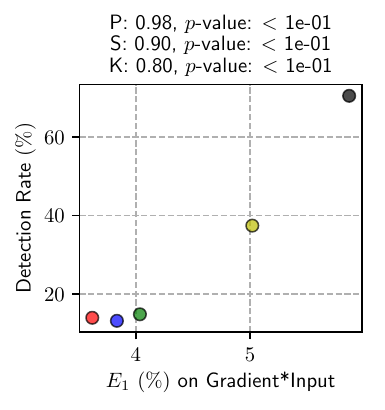}
    \includegraphics[width=0.446\textwidth,trim={.7cm 0 0 0},clip]{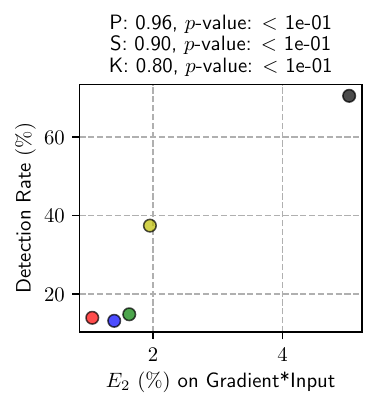}\\
    \includegraphics[width=0.5\textwidth,trim={0 0 0 0},clip]{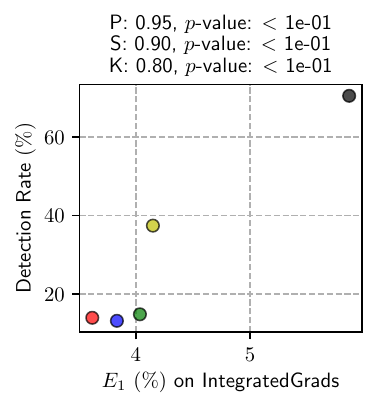}
    \includegraphics[width=0.446\textwidth,trim={.7cm 0 0 0},clip]{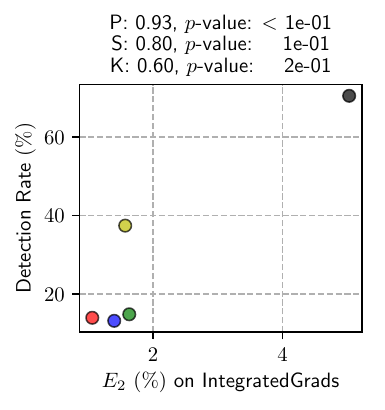}
    \caption{\DDrebin}
    \label{fig:er_scatter_drebin}
\end{subfigure}
\hfill
\begin{subfigure}[t]{0.49\textwidth}
    \centering
    \includegraphics[width=0.5\textwidth,trim={0 0 0 0},clip]{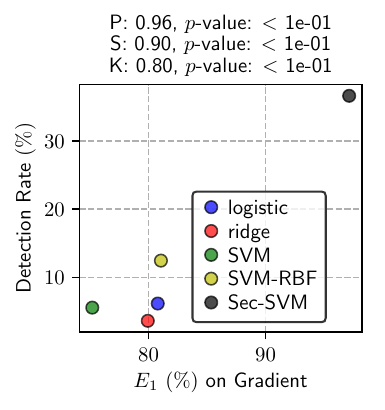}
    \includegraphics[width=0.446\textwidth,trim={.7cm 0 0 0},clip]{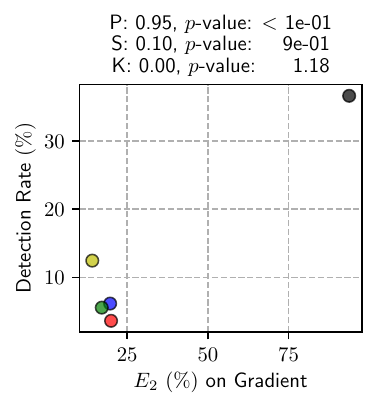}\\
    \includegraphics[width=0.5\textwidth,trim={0 0 0 0},clip]{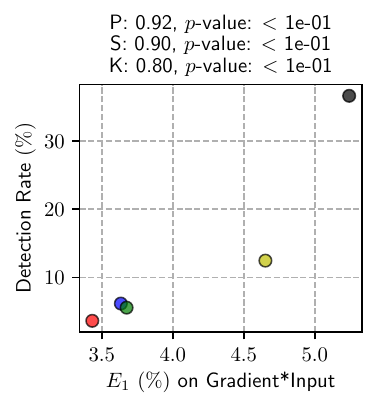}
    \includegraphics[width=0.446\textwidth,trim={.7cm 0 0 0},clip]{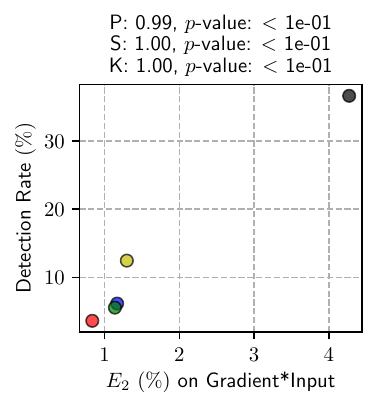}\\
    \includegraphics[width=0.5\textwidth,trim={0 0 0 0},clip]{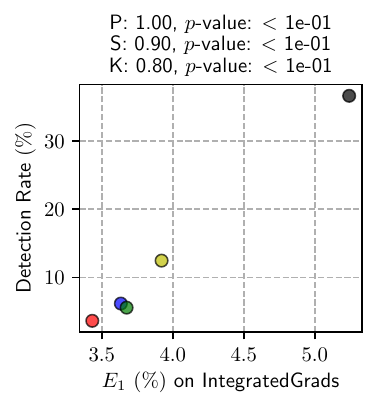}
    \includegraphics[width=0.446\textwidth,trim={.7cm 0 0 0},clip]{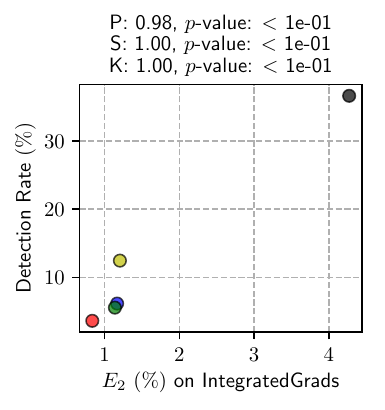}
    \caption{\DTesseract}
    \label{fig:er_scatter_parrot}
\end{subfigure}
    \caption{Evaluation of the evenness metrics $E_1$ (left) and $E_2$ (right) against the Detection Rate (FPR 1\%) for the different gradient-based explanation techniques}
    \label{fig:er_scatter}
\end{figure*}

We also inquire whether the connection between the evenness metrics and the detection performance of a classifier can provide a global assessment of its robustness.
Fig.~\ref{fig:er_scatter_drebin} and Fig.~\ref{fig:er_scatter_parrot} show the correlation between the explanation evenness and the mean detection rate under attack, calculated for $\varepsilon$ in the range $[1,50]$. Similarly to the previous tests, the uniformity metrics computed on the explanations from \gradin and \intgrad techniques present a significant connection to the detection rate, also witnessed by the $p$-values mostly under 0.01 for both datasets. Finally, the correlation with the \grad is again scarce, showing how this technique is not reliable to obtain information about the adversarial robustness of the tested classifiers to sparse evasion attacks.

\begin{table*}[t]
\begin{subtable}[t]{0.49\textwidth}
\centering
\begin{adjustbox}{width=\textwidth}
\begin{tabular}{r c cc | cc | cc}
 &  & \multicolumn{2}{c|}{\textbf{\grad}} & \multicolumn{2}{c|}{\textbf{\gradin}} & \multicolumn{2}{c}{\textbf{Int. Gradients}} \\ 
\cmidrule{3-8} 
 &  & $\evennx$ & $\evennxtwo$ & $\evennx$ & $\evennxtwo$ & $\evennx$ & $\evennxtwo$ \\ 
\toprule 
\textbf{\LOGI} & \makecell[tl]{P\\S\\K} &  &  & \makecell[tl]{
0.63, $<$1e-5\\0.66, $<$1e-5\\0.48, $<$1e-5} & \makecell[tl]{
0.71, $<$1e-5\\0.69, $<$1e-5\\0.51, $<$1e-5} & \makecell[tl]{
0.63, $<$1e-5\\0.66, $<$1e-5\\0.48, $<$1e-5} & \makecell[tl]{
0.71, $<$1e-5\\0.69, $<$1e-5\\0.51, $<$1e-5} \\ 
\midrule 
\textbf{\RIDGE} & \makecell[tl]{P\\S\\K} &  &  & \makecell[tl]{
0.47, $<$1e-5\\0.47, $<$1e-5\\0.33, $<$1e-5} & \makecell[tl]{
0.59, $<$1e-5\\0.59, $<$1e-5\\0.43, $<$1e-5} & \makecell[tl]{
0.47, $<$1e-5\\0.47, $<$1e-5\\0.33, $<$1e-5} & \makecell[tl]{
0.59, $<$1e-5\\0.59, $<$1e-5\\0.43, $<$1e-5} \\ 
\midrule 
\textbf{\SVM} & \makecell[tl]{P\\S\\K} &  &  & \makecell[tl]{
0.62, $<$1e-5\\0.65, $<$1e-5\\0.48, $<$1e-5} & \makecell[tl]{
0.67, $<$1e-5\\0.71, $<$1e-5\\0.54, $<$1e-5} & \makecell[tl]{
0.62, $<$1e-5\\0.65, $<$1e-5\\0.48, $<$1e-5} & \makecell[tl]{
0.67, $<$1e-5\\0.71, $<$1e-5\\0.54, $<$1e-5} \\ 
\midrule 
\textbf{\SVMRBF} & \makecell[tl]{P\\S\\K} & \makecell[tl]{
0.04, ~0.709\\0.41, $<$1e-4\\0.32, $<$1e-5} & \makecell[tl]{
0.68, $<$1e-5\\0.72, $<$1e-5\\0.54, $<$1e-5} & \makecell[tl]{
0.80, $<$1e-5\\0.94, $<$1e-5\\0.79, $<$1e-5} & \makecell[tl]{
0.77, $<$1e-5\\0.94, $<$1e-5\\0.80, $<$1e-5} & \makecell[tl]{
0.89, $<$1e-5\\0.94, $<$1e-5\\0.77, $<$1e-5} & \makecell[tl]{
0.91, $<$1e-5\\0.93, $<$1e-5\\0.77, $<$1e-5} \\ 
\midrule 
\textbf{\SECSVM} & \makecell[tl]{P\\S\\K} &  &  & \makecell[tl]{
0.77, $<$1e-5\\0.84, $<$1e-5\\0.66, $<$1e-5} & \makecell[tl]{
0.81, $<$1e-5\\0.87, $<$1e-5\\0.79, $<$1e-5} & \makecell[tl]{
0.77, $<$1e-5\\0.84, $<$1e-5\\0.66, $<$1e-5} & \makecell[tl]{
0.81, $<$1e-5\\0.87, $<$1e-5\\0.79, $<$1e-5} \\ 
\bottomrule 
\end{tabular}
\end{adjustbox}
\caption{\DDrebin}
\label{tab:mscore_correlation_drebin}
\end{subtable}
\hfill
\begin{subtable}[t]{0.49\textwidth}
\centering
\begin{adjustbox}{width=\textwidth}
\begin{tabular}{r c cc | cc | cc}
 &  & \multicolumn{2}{c|}{\textbf{\grad}} & \multicolumn{2}{c|}{\textbf{\gradin}} & \multicolumn{2}{c}{\textbf{Int. Gradients}} \\ 
\cmidrule{3-8} 
 &  & $\evennx$ & $\evennxtwo$ & $\evennx$ & $\evennxtwo$ & $\evennx$ & $\evennxtwo$ \\ 
\toprule 
\textbf{\LOGI} & \makecell[tl]{P\\S\\K} &  &  & \makecell[tl]{
0.40, $<$1e-4\\0.36, $<$1e-3\\0.25, $<$1e-3} & \makecell[tl]{
0.49, $<$1e-5\\0.41, $<$1e-4\\0.31, $<$1e-5} & \makecell[tl]{
0.40, $<$1e-4\\0.36, $<$1e-3\\0.25, $<$1e-3} & \makecell[tl]{
0.49, $<$1e-5\\0.41, $<$1e-4\\0.31, $<$1e-5} \\ 
\midrule 
\textbf{\RIDGE} & \makecell[tl]{P\\S\\K} &  &  & \makecell[tl]{
0.18, $<$1e-1\\0.26, $<$1e-2\\0.17, $<$1e-1} & \makecell[tl]{
0.10, $<$1e-1\\0.08, $<$1e-1\\0.07, $<$1e-1} & \makecell[tl]{
0.18, $<$1e-1\\0.26, $<$1e-2\\0.17, $<$1e-1} & \makecell[tl]{
0.10, $<$1e-1\\0.08, $<$1e-1\\0.07, $<$1e-1} \\ 
\midrule 
\textbf{\SVM} & \makecell[tl]{P\\S\\K} &  &  & \makecell[tl]{
0.46, $<$1e-5\\0.37, $<$1e-3\\0.26, $<$1e-4} & \makecell[tl]{
0.31, $<$1e-2\\0.24, $<$1e-1\\0.17, $<$1e-1} & \makecell[tl]{
0.46, $<$1e-5\\0.37, $<$1e-3\\0.26, $<$1e-4} & \makecell[tl]{
0.31, $<$1e-2\\0.24, $<$1e-1\\0.17, $<$1e-1} \\ 
\midrule 
\textbf{\SVMRBF} & \makecell[tl]{P\\S\\K} & \makecell[tl]{
-0.78, $<$1e-5\\-0.64, $<$1e-5\\-0.45, $<$1e-5} & \makecell[tl]{
-0.58, $<$1e-5\\-0.52, $<$1e-5\\-0.35, $<$1e-5} & \makecell[tl]{
0.88, $<$1e-5\\0.85, $<$1e-5\\0.67, $<$1e-5} & \makecell[tl]{
0.66, $<$1e-5\\0.56, $<$1e-5\\0.41, $<$1e-5} & \makecell[tl]{
0.88, $<$1e-5\\0.79, $<$1e-5\\0.61, $<$1e-5} & \makecell[tl]{
0.54, $<$1e-5\\0.45, $<$1e-5\\0.31, $<$1e-5} \\ 
\midrule 
\textbf{\SECSVM} & \makecell[tl]{P\\S\\K} &  &  & \makecell[tl]{
0.61, $<$1e-5\\0.42, $<$1e-4\\0.30, $<$1e-4} & \makecell[tl]{
0.66, $<$1e-5\\0.48, $<$1e-5\\0.35, $<$1e-5} & \makecell[tl]{
0.61, $<$1e-5\\0.42, $<$1e-4\\0.30, $<$1e-4} & \makecell[tl]{
0.66, $<$1e-5\\0.48, $<$1e-5\\0.35, $<$1e-5} \\ 
\bottomrule 
\end{tabular}
\end{adjustbox}
\caption{\DTesseract}
\label{tab:mscore_correlation_parrot}
\end{subtable}
\caption{Correlation between the adversarial robustness metric $\set R$ and the evenness metrics $\evennx$ and $\evennxtwo$. Pearson (P), Spearman Rank (S), Kendall’s Tau (K) coefficients along with corresponding $p$-values. The linear classifiers lack a correlation value since the evenness is constant (being the gradient constant as well), thus resulting in a not defined correlation}
\label{tab:mscore_correlation}
\end{table*}

\section{Related Work}
\label{sect:relwork}
In this section, we provide an overview of the literature on Android Malware Detection systems (Sect. \ref{subsect:maldet}), on the techniques to craft powerful adversarial attacks against them (Sect. \ref{subsect:advmal}), and, finally, on the approaches to explain their decisions (Sect. \ref{subsect:relexplain}).

\subsection{Android Malware Detection}
\label{subsect:maldet}

The detection of Android malware attacks has been addressed over the years through works leveraging static, dynamic, or hybrid analyses.

Arzt et al.~\cite{Arzt2013-PLDI} proposed FlowDroid, a security tool that performs static taint analysis within the single components of Android applications. Feng et al.~\cite{Feng2014-FSE} proposed Apposcopy, a detection tool that combines static taint analysis and intent flow monitoring to produce a signature for applications. Tam et al.~\cite{tam15-ndss} proposed CopperDroid, a dynamic analyzer that aims to identify suspicious high-level behaviors of malicious Android applications. 
More recently, \textsc{MaMaDroid} by Mariconti et al.~\cite{mariconti-ndss17} employs Markov chains to model sequences of API calls. Chen et al.~\cite{Chen2021} converted app opcodes to an image-like structure in order to perform data augmentation through a Generative Adversarial Network (GAN), while the works by Mahindru et al. focused on assessing effective feature selection, mainly considering the usage of APIs and permissions as features~\cite{Mahindru2021,Mahindru2021a}.
Moreover, different works in the literature target specific types of attacks, such as botnets \cite{Hijawi2021} or ransomware samples \cite{maiorca17-sac,chen18-tifs,scalas19-cose}.

An interesting aspect to underline is that most of the feature sets used in previous work~---~the earliest as well as the newest ones~---~include information from Android APIs~\cite{Li2021,rieck14-drebin,maiorca17-sac,mariconti-ndss17,Mahindru2021,AafDuYin13,Chen2018,scalas19-cose}. According to Zhang et al.~\cite{Zhang2020}, although Android malware evolves over time, many semantics are still the same or similar, and can be caught by identifying the relations between the different APIs. 
In particular, several works other than Drebin \cite{rieck14-drebin} inspect the usage of certain APIs \cite{AafDuYin13,Mahindru2021} or the number of API calls \cite{Chen2018,scalas19-cose}, which typically implies the design of \emph{sparse} feature vectors for ML-based detectors. This approach is often valid for other Android components (\eg, the usage of permissions). Hence, suggesting that our analysis is likely to be relevant and applicable to many other detectors in the literature.

\subsection{Adversarial attacks}
\label{subsect:advmal}

According to a recent survey by Biggio et al. \cite{biggio18-pr}, several works questioned the security of machine learning since $2004$. Two pioneering works were proposed by Dalvi et al. \cite{dalvi04-kdd} in 2004 and by Lowd and Meek \cite{lowd05-kdd} in 2005. Those works, considering linear classifiers employed to perform spam filtering, demonstrated that an attacker could easily deceive the classifier at test time (\emph{evasion} attacks) by performing a limited amount of carefully-crafted changes to an email. 
Subsequent works \cite{barreno06-asiaccs,barreno10,biggio14-tkde} proposed attacker models and frameworks that are still used to study the security of learning-based systems also against training-time (\emph{poisoning}) attacks. 
The first gradient-based poisoning \cite{biggio12-icml} and evasion \cite{biggio13-ecml} attacks were proposed by Biggio et al. respectively in 2012 and 2013. Notably, in \cite{biggio13-ecml} the authors also introduced two important concepts that are still heavily used in the adversarial field, namely \emph{high-confidence} adversarial examples and the use of a \emph{surrogate} model. This work anticipated the discovery of the so-called \emph{adversarial examples} against deep networks \cite{szegedy14-iclr,goodfellow15-iclr}.

The vulnerability to evasion attacks was then studied especially on learning systems designed to detect malware samples (for example, on PDF files \cite{maiorca19-csur,srndic14}), thus raising serious concerns about their usability under adversarial environments.
In particular, for Android malware detectors, Demontis et al. \cite{demontis17-tdsc} demonstrated that linear models trained on the (static) features extracted by \Drebin can be easily evaded by performing a fine-grained injection of information (a more advanced injection approach that directly operates on the Dalvik bytecode has been proposed by Yang et al. \cite{yang17-acsac}) by employing gradient descent-based approaches. Grosse et al. \cite{grosse17-esorics} have also attained a significant evasion rate on a neural network trained with the \Drebin feature set.
Although the adversarial robustness of other Android detectors aside from \cite{rieck14-drebin} was not fully explored, it is evident that employing information that can be easily injected or modified may increase the probability of the attacker to attain successful evasion.

However, as discussed in Sect. \ref{sect:advandroid}, adding or removing (modifying) parts of a sample to create adversarial attacks is an apparently-straightforward operation. In practise, the real-world feasibility and the constraints of these operations should be carefully evaluated. Only recently, research efforts were spent into investigating \emph{problem-space attacks}, focusing on the generation of real evasive samples \cite{pierazzi2020intriguing, cara2020feasibility}. Given that in the software domain there is no clear inverse mapping to the feature space, unlike in computer vision for example (so that the app's semantics are correctly preserved), this research direction remained underexplored for many years.

\subsection{Explainability}
\label{subsect:relexplain}

Consequently to the rise of black-box models in the last decade, explainability became a hot research topic. It can be leveraged to achieve multiple goals, from justifying each prediction (the \emph{right of explanation} required by the European General Data Protection Regulation (GDPR) \cite{goodman16-gdpr}) to discovering new knowledge and causal relations.
Explainability became increasingly popular in security as well, as providing a proper explanation of predictions can help to secure the systems against adversarial attacks.

Several approaches for interpretability have been proposed, with a particular attention to \emph{post-hoc} explanations for black-box models.
In 2016, Ribeiro et al. \cite{ribeiro16} proposed LIME, a model-agnostic technique that provides local explanations by generating small perturbations of the input sample, thus obtaining the explanations from a linear model fitted on the perturbed space. Lundberg and Lee \cite{Lundberg2017} unified different techniques, including LIME, under the name of SHAP, by leveraging cooperative game theory results to identify theoretically-sound explanation methods and provide feature importance for each prediction. More recently, Lundberg et al.~\cite{Lundberg2020} improved this method for tree-based models, including those based on multiple trees, by mantaining the desidered properties for local explanations and enabling faithful \emph{global} understanding of the models as well.
The work by Koh and Liang \cite{koh17-icml} showed that using a gradient-based technique called \emph{influence functions}, which is well known in the field of robust statistics, it is possible to associate each input sample to the training samples (\emph{prototypes}) that are most responsible for its prediction. The theory behind the techniques proposed by the authors holds only for classifiers with differentiable loss functions. However, the authors empirically showed that their technique provides sensible prototypes also for classifiers with not-differentiable losses if computed on a smoothed counterpart.

Another interesting venue is the generation of high-level \emph{concepts} rather than feature attributions. In this sense, Kim et al.~\cite{Kim2018-icml} proposed a technique that introduces the notion of \textit{Concept Activation Vectors} (CAVs), which evaluate the sensitivity of the models to user-defined examples defining particular concepts. Koh et al.~\cite{koh20-icml} focused instead on guiding models to learn concepts at training time; such concepts are then used to predict the target samples.

Notably, as recent work started explaining malware detectors through some of the above-described techniques \cite{melis2018explaining, Scalas2020ICPR}, Warnecke et al.~\cite{Warnecke2020} discussed general and security-specific criteria to evaluate explanation methods in different security domains. Moreover,
Guo et al.~\cite{Guo2018} proposed LEMNA, a method specifically designed for security tasks, \ie, that is optimized for RNN and MLP networks, and that highlights the feature dependence (\eg, for binary code analysis). 

Finally, a few recent works proposed to leverage explanations for both generating and detecting attack samples.
Rosenberg et al. \cite{rosenberg2020generating} obtained from explainability algorithms the most relevant features for a malware classification task, so that those can be the first to be modified in order to generate an effective adversarial attack. Starting from the same idea, Fidel et al.~\cite{fidel2020explainability} proposed a highly accurate detector of adversarial examples, based on the SHAP values computed for the internal layers of a DNN classifier. Also, Dombrowski et al. showed how saliency maps can be manipulated arbitrarily by applying perturbations to the input, while keeping the model’s output approximately constant \cite{dombrowski2019explanations}. This is a worst case scenario where not only the prediction of the system is wrong (the perturbed malicious point evades detection), but also the explanation that might have been used to identify the vulnerability is compromised.

\section{Conclusions and Future Work} \label{sect:conclusions}

In this paper, we empirically evaluate the correlation between multiple gradient-based explanation techniques and the \emph{adversarial robustness} of different linear and non-linear classifiers, trained on two popular Android applications datasets (\DDrebin and \DTesseract), against sparse evasion attacks. To this end, we leverage two synthetic measures of the \emph{explanation evenness}, which main advantage is not requiring any computationally-expensive attack simulations. Thus, they may be used by system designers and engineers to choose, among a plethora of different models, the one that is most resilient against sparse attacks. Our experiments show that a strong connection exists between the evenness of explanations and the adversarial robustness. This correlation is stronger when advanced explanation techniques such as \gradin and \intgrad are used, while considering the simple \grad does not provide reliable information about the robustness of such classifiers.

In the future, we plan to extend our study to other malware detectors as well as other application domains. Moreover, as the proposed metrics may be used to estimate the robustness only against sparse evasion attacks in a boolean feature space, such as the one of \Drebin, an interesting research direction would be to devise a similar measure that can be used to estimate the robustness of detectors working in continuous, dense feature spaces, and when the attack is subjected to different application constraints. Also, it could be interesting to assess if our vulnerability measures can be successfully applied when the attacker does not know the classifier parameters or when the model is not differentiable; in that case, a surrogate classifier would be used to explain the original unknown model function. 

Finally, another interesting research avenue is to modify the objective functions used to train the considered machine learning models by adding to them a penalty which is inversely proportional to the proposed evenness metrics, in order to enforce the classifier to learn more evenly distributed relevance scores and, consequently, the model robustness.

\begin{acknowledgements}
This work has been partly supported by the PRIN 2017 project RexLearn (grant no.~2017TWNMH2), funded by the Italian Ministry of Education, University and Research, and by BMK, BMDW, and the Province of Upper Austria in the frame of the COMET Programme managed by FFG in the COMET Module S3AI.
\end{acknowledgements}

\section*{Data availability}
The datasets generated during and/or analysed during the current study are available in the \texttt{Androzoo} repository, \url{https://androzoo.uni.lu/}, and upon request at \url{https://www.sec.tu-bs.de/~danarp/drebin/}.

\section*{Conflict of interest}
The authors declare that they have no conflict of interest.

\bibliographystyle{spmpsci}      
\bibliography{add-refs,bibDB,malwDB}

\end{document}